%% file: main2.tex
\documentclass[letterpaper, 10 pt, conference]{ieeeconf}  

\IEEEoverridecommandlockouts                              

\title{UST: Unifying Spatio-Temporal Context for Trajectory Prediction in Autonomous Driving}

\author{Hao He$^{1}$, Hengchen Dai$^{1}$, Naiyan Wang$^{1}$
\thanks{$^{1}$Hao He, Hengchen Dai and Naiyan Wang are with
TuSimple, Beijing. China. \{{\tt\small hugohaohe, dhcisyx, winsty\}@gmail.com} }
}

\usepackage{times}
\usepackage{epsfig}
\usepackage{graphicx}
\usepackage{amsmath}
\usepackage{amssymb}
\usepackage{epigraph}
\usepackage[pagebackref=true,letterpaper=true,colorlinks,bookmarks=false]{hyperref}
\usepackage{algorithm}

\usepackage{listings}
\usepackage{etoolbox}

\usepackage{booktabs}
\usepackage{custom_global}
\usepackage{multirow}
\usepackage{multicol}
\usepackage{array}
\usepackage{subfigure}

\label{User-Defined-Begin}

\def\p{{\mathbf p}}
\def\X{{\mathbf X}}
\def\F{{\mathbf F}}
\def\H{{\mathbf H}}
\def\x{{\mathbf x}}

\makeatletter
\AfterEndEnvironment{algorithm}{\let\@algcomment\relax}
\AtEndEnvironment{algorithm}{\kern2pt\hrule\relax\vskip3pt\@algcomment}
\let\@algcomment\relax
\newcommand\algcomment[1]{\def\@algcomment{\footnotesize#1}}
\renewcommand\fs@ruled{\def\@fs@cfont{\bfseries}\let\@fs@capt\floatc@ruled
  \def\@fs@pre{\hrule height.8pt depth0pt \kern2pt}%
  \def\@fs@post{}%
  \def\@fs@mid{\kern2pt\hrule\kern2pt}%
  \let\@fs@iftopcapt\iftrue}
\makeatother

\newcolumntype{+}{>{\global\let\currentrowstyle\relax}}
\newcolumntype{Y}{>{\currentrowstyle}}
\newcommand{\rowstyle}[1]{\gdef\currentrowstyle{#1}%
	#1\ignorespaces
}
\label{User-Defined-End}

\begin{document}

\maketitle
\thispagestyle{empty}
\pagestyle{empty}

\input{content/abstract}

\input{content/introduction}

\input{content/related_work}

\input{content/method}

\input{content/experiment}

\input{content/result}

\input{content/conclusion}

\bibliographystyle{IEEEtran}
\bibliography{egbib}

\end{document}

%% file: content/abstract.tex
\begin{abstract}
    Trajectory prediction has always been a challenging  problem for autonomous driving, since it needs to infer the latent intention from the behaviors and interactions from traffic participants. This problem is intrinsically hard, because each participant may behave differently under different environments and interactions. 
    This key is to effectively model the interlaced influence from both spatial context and temporal context.
    Existing work usually encodes these two types of context separately, which would lead to inferior modeling of the scenarios.
    In this paper, we first propose a unified approach to treat time and space dimensions equally for modeling spatio-temporal context. The proposed module is simple and easy to implement within several lines of codes. In contrast to existing methods which heavily rely on recurrent neural network for temporal context and hand-crafted structure for spatial context, our method could automatically partition the spatio-temporal space to adapt the data.
    Lastly, we test our proposed framework on two recently proposed trajectory prediction dataset ApolloScape and Argoverse. 
    We show that the proposed method substantially outperforms the previous state-of-the-art methods while maintaining its simplicity.
    These encouraging results further validate the superiority of our approach.
\end{abstract}

%% file: content/introduction.tex
\section{Introduction}
With the great development of deep learning techniques in recent years, the perception system equipped in autonomous driving system has been significantly advanced. However, the difficulty of another equally important task, predicting the future trajectories for traffic participants in real-world scenarios is still underestimated, since trajectory prediction task requires understanding the latent intention from the behaviors and interactions of participants.

\begin{figure}[!t]
    \centerline{
        \subfigure[Illustration]{\includegraphics[width=0.17\textwidth]{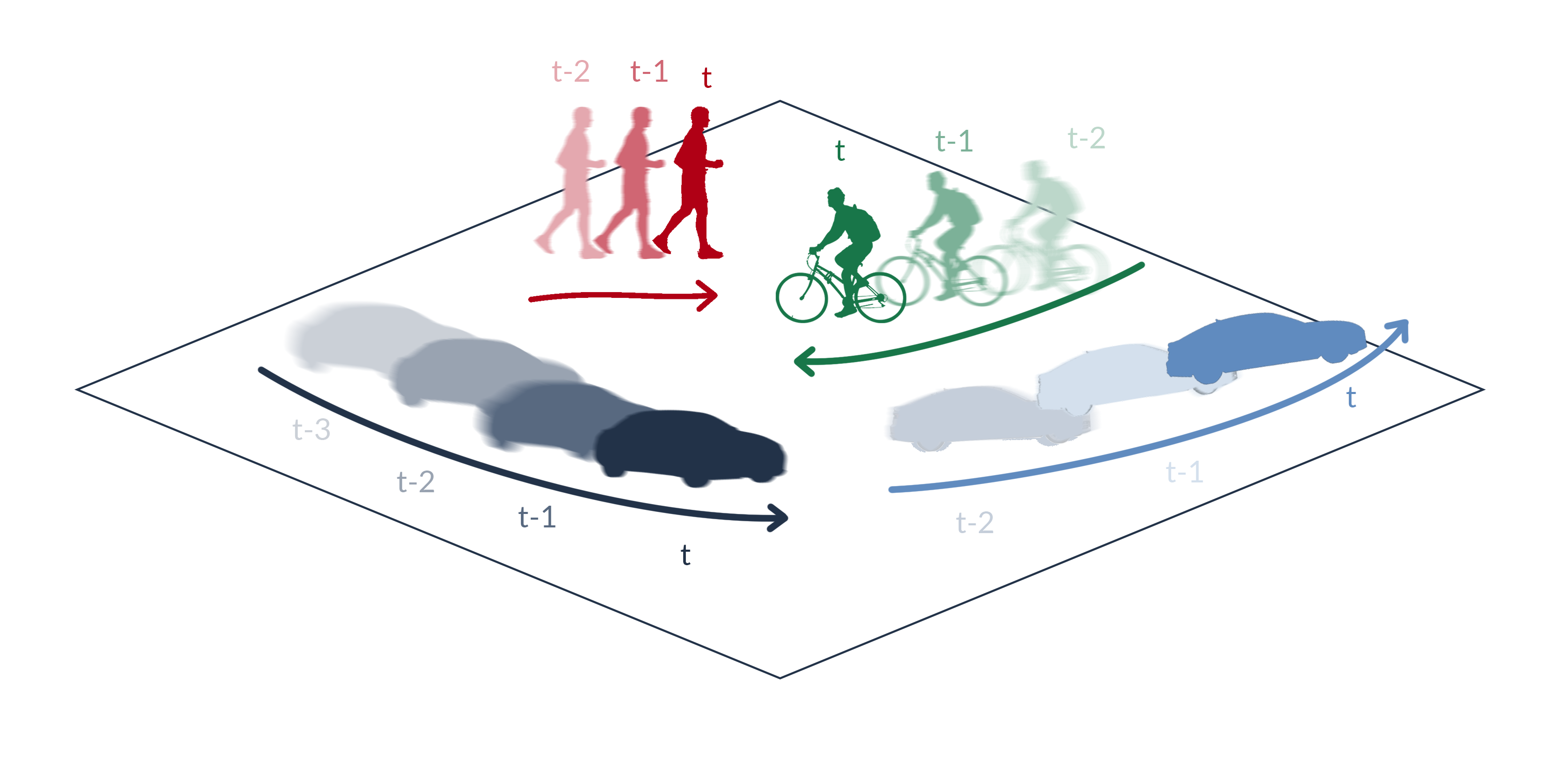} \label{fig:illustration}}
    \subfigure[Traditional Representation]{\includegraphics[width=0.17\textwidth]{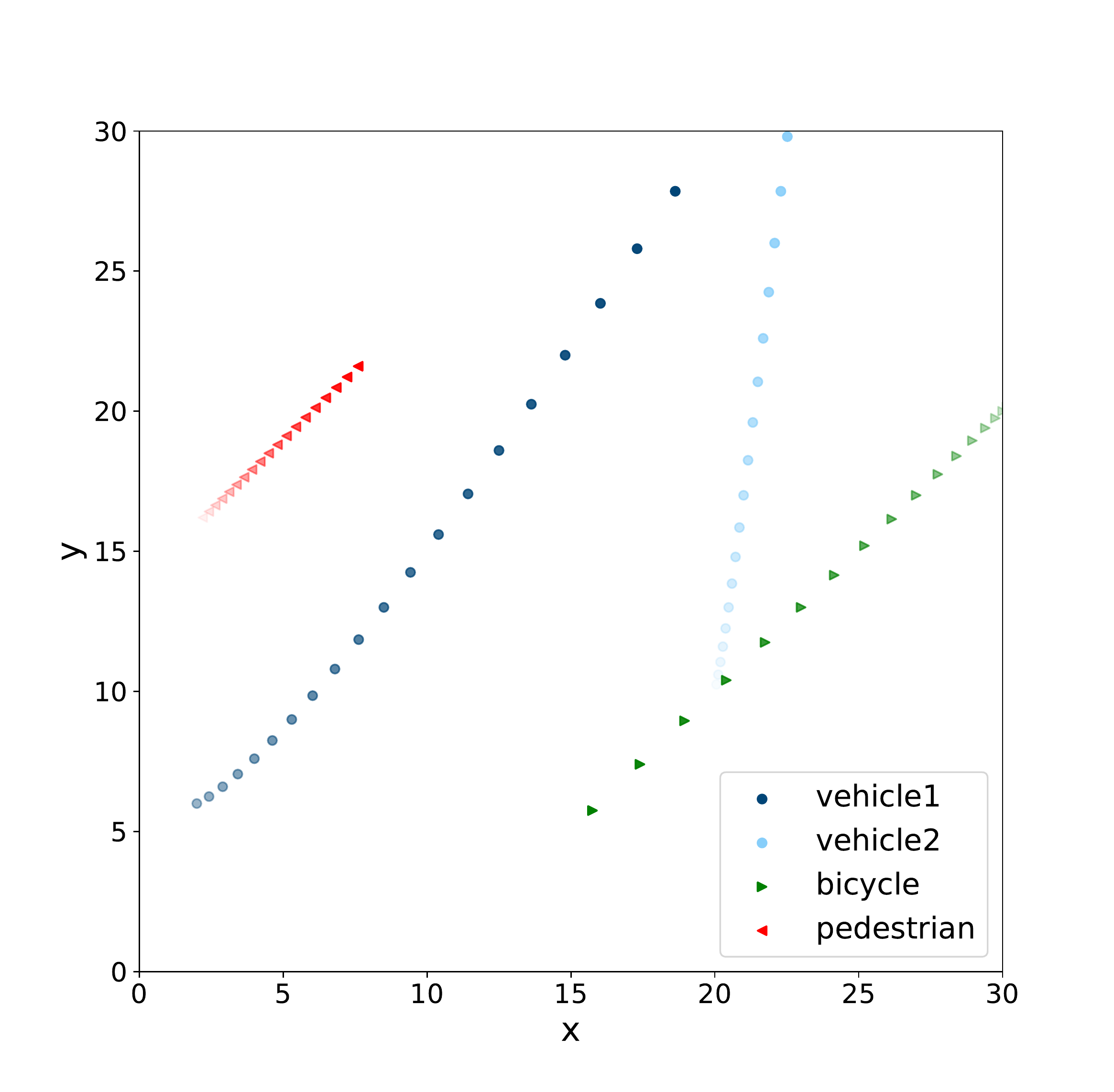} \label{fig:input_representation_2d}}
    \subfigure[Unified 3D Representation]{\includegraphics[width=0.17\textwidth]{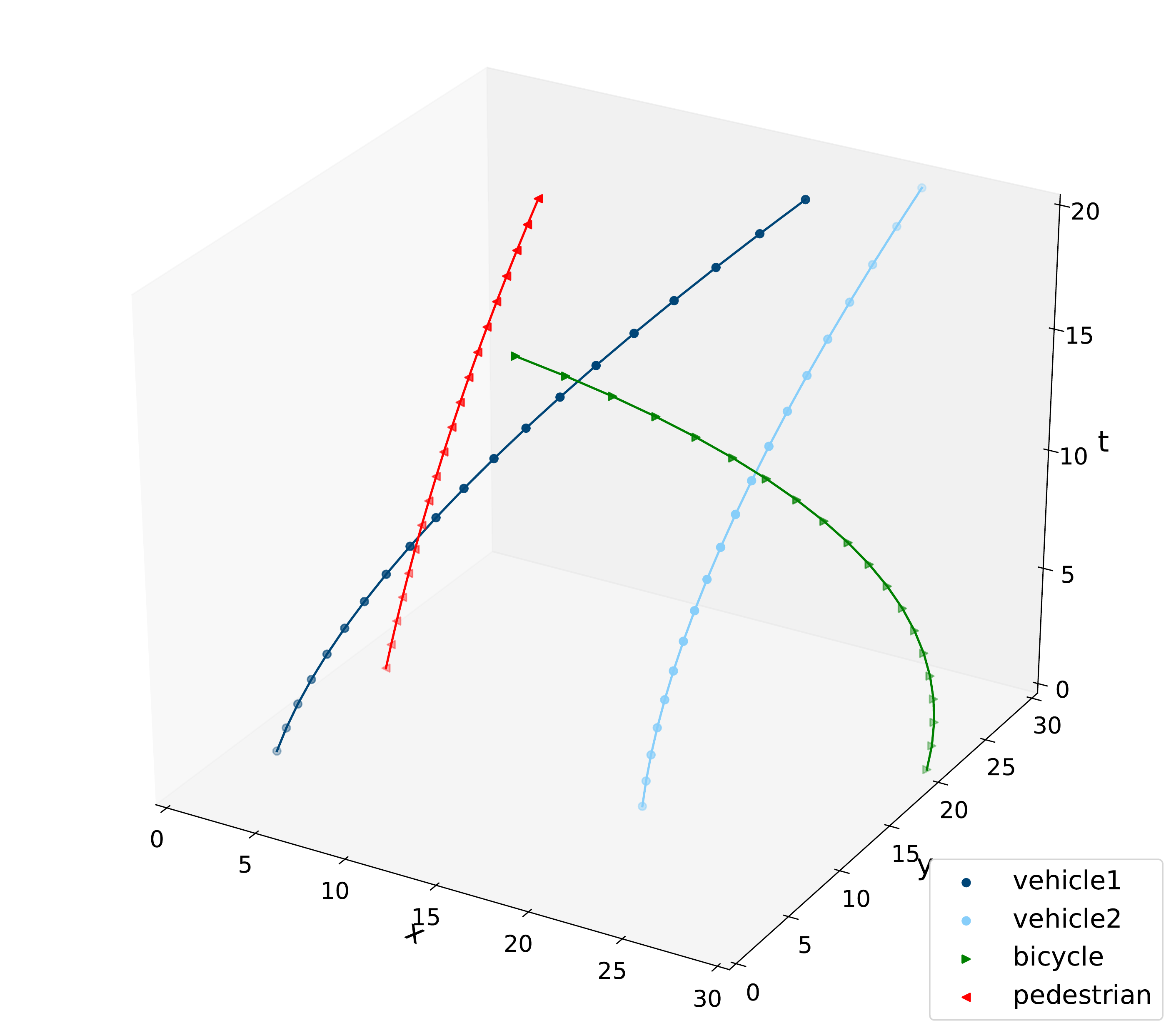} \label{fig:input_representation_3d}}
    }
    \caption{
        Illustration and representations of the trajectory prediction task.
        Blue, green, red colors show trajectories for vehicles, bicycles, pedestrians respectively.
        (b) shows the common representation, which represents the surrounding agents as sequences of positions in 2D spatial space.
        (c) shows our proposed trajectory representation in a unified spatio-temporal space.
    }
    \label{fig:input_representation}
\end{figure}

The key to this challenging task is to model the \textit{complicated spatio-temporal social context} and tolerate \textit{the imperfect output from perception systems}. 
For the first task, the difficulty lies in that we can only infer the intention from indirect observations. In a certain circumstance, different drivers or pedestrians may make distinct decisions. On the other hand, even for a certain driver or pedestrian, its behavior may be easily affected by diverse interactions at different places and different time (see Fig.~\ref{fig:illustration}).
For the second challenge, we cannot expect an oracle perception system due to occlusion and limited range of sensors. Common mistakes include trajectory interruption, unstable speed estimation, etc.
Existing works always make the strong assumption that each neighbor has a fix length trajectory and highly depend on accurate speed feature. Some works even eliminate those agents who have incomplete history\cite{gupta2018social, deo2018convolutional}. However, these are just the common scenarios we meet in real autonomous driving systems.

Most learning-based trajectory prediction methods can be categorized into the following encoder-decoder framework: they usually consist of \textit{spatio-temporal context encoder} and \textit{future trajectory decoder}.
In the first module, we want to utilize all the available history information to model the intention of agents.
The information we can use include the history trajectory of one single agent (temporal context) and the interaction between agents in one single step (spatial context). Other meta-data such as maps or traffic rules can also be incorporated. While in the second module, given the encoded context, we need to generate the future trajectory of the agent. The future trajectory can be represented either by a deterministic path~\cite{zhang2019sr, li2019grip}, a path with uncertainty estimation~\cite{alahi2016social, vemula2018social} or several sampling paths~\cite{gupta2018social, zhao2019multi}.


Most existing spatio-temporal context encoders encode spatial context and temporal context separately.
They typically extract each agent's temporal context with a recurrent neural network (RNN).
After that, a spatial context extractor (e.g., convolution~\cite{deo2018convolutional}, attention~\cite{messaoud2019non}, pooling~\cite{gupta2018social}, graph network~\cite{kosaraju2019social}) will be applied to further aggregate the spatial context.
The spatial information between agents is considered at the last time step, thus the interaction among agents in previous time steps are all abandoned.
To improve upon these methods, recent works~\cite{vemula2018social, huang2019stgat} have proposed to leverage spatio-temporal graph. The basic idea of these methods is to extracting spatial context at each time step, and then fuse it into individual temporal features for next time aggregation.
However, they still extract spatio-temporal context in this cascade style (first temporal, then spatial) essentially.
On one hand, extracting spatial context at each time step can be time-consuming (reported in~\cite{gupta2018social} as 16x slower); on the other hand, it cannot model complex spatial context across different time step.
There is also an intrinsic problem involved with RNN-based temporal context extractor.
When existing missing data, either eliminating the whole history or padding with interpolations is necessary.
All these drawbacks call for a unified spatio-temporal modeling method for trajectory prediction.

In this paper, we propose a novel method to address the above challenges by Unifying the Spatial and Temporal context ~(\pmethodb) into one single representation.
The core idea of \pmethodb~is to jointly represent the spatio-temporal context in a higher dimensional space, which does not distinguish these two concepts explicitly. Then the status of a certain agent at a certain time is represented as a point in this space. (See Fig.~\ref{fig:input_representation_3d} for illustration.) Other meta-data can also be easily incorporated.
As a result, the spatio-temporal context becomes the distribution of these unordered points in the new space.
Next, inspired by the classical work PointNet~\cite{qi2017pointnet}, we devise a novel encoder structure for joint context modeling.
Based on the extracted context, various off-the-shelf decoders can be used to incorporate different purposes from subsequent modules.
To summarize, our main contributions are as follows:
\begin{itemize}
    \item We first propose to unify 2D location and discrete time space into one single 3D space. We treat them equally. 
    \item Based on this representation, we devise a simple and effective network to encode the spatio-temporal context. The method can be easily implemented within ten lines of codes.
    \item Experiments on three major trajectory prediction datasets demonstrates the effectiveness of UST.
\end{itemize}

%% file: content/related_work.tex
\section{Related Work}
In this section, we will give a brief review of the history and recent researches about trajectory prediction problem.

The idea of predicting 2D future trajectories using spatio-temporal context could date back to the work of Helbing and Molnar~\cite{helbing1995social}, who proposed a model with attractive and repulsive \textit{Soical Force}.
It has achieved remarkable successes in robotics\cite{luber2010people, ferrer2013robot} and activity understanding\cite{yamaguchi2011you, choi2013understanding}. 
Recently, many researchers extended these interaction-aware trajectory prediction methods to a variety of areas with advanced deep learning techniques\cite{alahi2016social, gupta2018social, zhang2019sr, zhao2019multi}, including pedestrians trajectory prediction\cite{alahi2016social}, vehicles trajectory prediction\cite{deo2018convolutional}, intention prediction\cite{ding2019predicting} and heterogeneous prediction\cite{ma2019trafficpredict}.
Despite they are dealing with various tasks with different approaches, we summarize most deep learning based methods into the \textit{spatio-temporal context encoder} and \textit{future trajectory decoder} framework.

\paragraph{Spatio-temporal Context Encoder} 
The raw data of trajectory prediction task consists of both spatial and temporal information and various metadata. They cannot be directly processed by off-the-shelf neural networks.
Popular representation forms include sequential points-based~\cite{gupta2018social, vemula2018social, huang2019stgat}, occupy gird-based~\cite{kim2017probabilistic, deo2018convolutional}, displacement volume~\cite{yi2016pedestrian, li2019way} and rasterized image~\cite{cui2019multimodal, djuric2018motion}.
Directly representing trajectory as a sequence of 2D location is the most natural way, however its unstructured characteristics make context extraction hard.
Although all other representations preserve the structural information, their structures are hand-crafted thus need deliberately tuning, which makes the generalization across different scenarios infeasible. Moreover, these hand-crafted structures are sensitive to the quality of perception output. If the quality of the output of upstream modules changes, the structure may need to be redesigned. Consequently, how to adaptively build the structure from data still remains an open problem.

Based on these representations, researchers have devised various networks to extract the spatio-temporal context.
However, all existing work extract the temporal context and spatial context individually, and aggregate them in a cascade way.
For the temporal context, Recurrent Neural Network (RNN) is the most widely used.
There are also some other algorithms that use Convolution Neural Network (CNN) to encode temporal context\cite{nikhil2018convolutional, li2019grip}.
For the spatial context (a.k.a ``social context''), its extaction remains to a hot research topic in recent years. Various methods including convolution\cite{deo2018convolutional}, pooling\cite{alahi2016social,gupta2018social}, attention network\cite{vemula2018social}, graph neural network\cite{huang2019stgat}, relation network\cite{choi2019looking} have been used in past few years. 
Instead of manually building the spatial partition and then encoding the temporal context and spatial context separately, our proposed \pmethodb~unifies the spatio-temporal space and learn to partition this joint space end-to-end.

\textbf{Future Trajectory Generation} 
The task of the generator or decoder is to generate the future position of the target agent based on both the ego information and the encoded context information. The most common one is to use a recurrent neural network (RNN) to regress the future trajectory directly\cite{huang2019stgat, zhang2019sr}.
There are also other variants based on RNN to accommodate other demands from downstream modules.
To name a few, Gupta \textit{et al.}\cite{gupta2018social} incorporated noise to the RNN decoder and train it with a discriminator and variety loss to generate diverse socially acceptable trajectories.
Deo \textit{et al.} \cite{deo2018convolutional} tried to classify maneuvers firstly then use the classification result to construct a Gaussian mixture model for multi-modal trajectory prediction.
Chai \textit{et al.} \cite{chai2019multipath} classified trajectories to the pre-clustered anchors instead of predefined maneuvers.
In this work, decoder is not our focus. Thus we directly utilize the off-the-shelf decoders in our proposed method.


%% file: content/method.tex
\section{Method}


This section gives details of our proposed method.
Subsection \ref{Problem Definition} formalizes the trajectory prediction problem mathematically.
Subsection \ref{encoder} details how to encode the spatial-temporal context in a unified framework.
Subsection \ref{decoder} elaborates how we use the encoded spatio-temporal context representation to generate future trajectories.

\subsection{Problem Definition}\label{Problem Definition}
Let's assume that there are $N$ agents and $T$ time steps in total, and $\x_n^t \in \mathbb{R}^2$ denotes the raw 2D spatial location of agent $n$ at time step $t$ with respect to a predefined reference frame.
We can further formulate the history of one single agent $n$ as $\X_n \doteq \{\H_n, \F_n\}$, where $\H_n \doteq \{x_n^1, x_n^2, \cdots, x_n^t\}$ refers to history locations of agent $n$ starting from time $1$ to time $t$, while $\F_n \doteq \{x_n^{t+1}, x_n^{t+2}, \cdots, x_n^T\}$ refers to future locations of agent $n$ starting from time $t+1$ to time $T$.
$\mathcal{I}$ is the additional information such as traffic-agent type, map information, taillight, heading, etc.
The goal of trajectory prediction is to utilize all $\{\H_n\}_{n=1}^N$ and $\mathcal{I}$ to accurately predict $\{\F_n\}_{n=1}^N$.

\subsection{Spatio-temporal Context Encoder}\label{encoder}
In this subsection, we elaborate details of our spatio-temporal context encoder. We first introduce the input representation of our method, then followed by the encoder structure. 

\subsubsection{Input Representation}\label{method_input_representation}
For the sake of simplicity and flexibility, we design spatio-temporal point sets to represent the raw input (sequence of locations of multiple agents) in the following form:
\begin{equation}
\begin{aligned}
    &\p_n^t = \{x_n^t, v_n^t, \phi_n, t, \mathrm{id}\}, \\
    &\mathcal{P} = \{p_n^t \mid \forall t \in \{1,...,T\},\forall n \in \{1,...,N\} \} 
\end{aligned}
\end{equation}
where $x_n^t, v_n^t$ are 2D locations and 2D velocity of agent $n$ in current reference frame, $\phi_n$ denotes some intrinsic properties of agent $n$ such as type.
$t \in \{1,...,T\}$ refers to the time step. And $\mathrm{id}$ is a binary number, 0 is for neighbor agents, and 1 for the target agent to be predicted. We don't distinguish different neighborhood agents, and treat them equally.

After applying the above input representation module, we can treat the snapshot of status of agent $n$ at time step $t$ as a single point with metadata in a 3D space spanned by 2D location and time. An intuitive illustration is depicted in Fig.~\ref{fig:input_representation_3d}.
Note that all the information we need to model the social interaction of the target agent $n$ is presented in $\mathcal{P}$. One desired property of this representation is that it is invariant to the order of each $\p_n^t$ in it and robust to missing data.
By this uniform representation, we unify space and time into one representation, which eases the subsequent context modeling task.

\subsubsection{Unified Spatio-Temporal Context Extraction}
To deal with such unordered and variable length data, the structure and operations of this feature extractor should be deliberately designed to fit the nature of the data. Fortunately, the same challenge has been met in the area of point cloud processing. Inspired by PointNet~\cite{qi2017pointnet}, we propose the following two key components for context extraction. 
\paragraph{Embedding} The goal of this embedding network $\phi$ is to map $p_n^t$ into a hidden representation $e_n^t$, in which the spatial context and temporal context are unified:
\begin{equation}
    \begin{aligned}
        e_n^t &= \phi(p_n^t; W_{e}). 
    \end{aligned}
\end{equation}


In our implementation, $\phi$ is implemented by a multiple layer perceptron (MLP).
$W_{e}$ is the embedding weight.
Batch normalization is applied over all layers with ReLU activation functions.
Although there are several works that also apply MLP on time series prediction\cite{koskela1996time}, they treat the whole time series as a fixed-length chronological data instead of a set of permutation invariant snapshots.


\paragraph{Permutation Invariant Aggregator} 
After embedding each $\p_n^t$, we need a permutation invariant aggregator to form the global context feature. ~\cite{qi2017pointnet} has shown that simple pooling operator is capable for the task. It is the simplest symmetric function that enjoys this property.
By default, we use max pooling as the aggregator.

\begin{figure*}[!htb]
    \centerline{
        \subfigure[Partition for left front and right front area]{\includegraphics[width=0.3\textwidth]{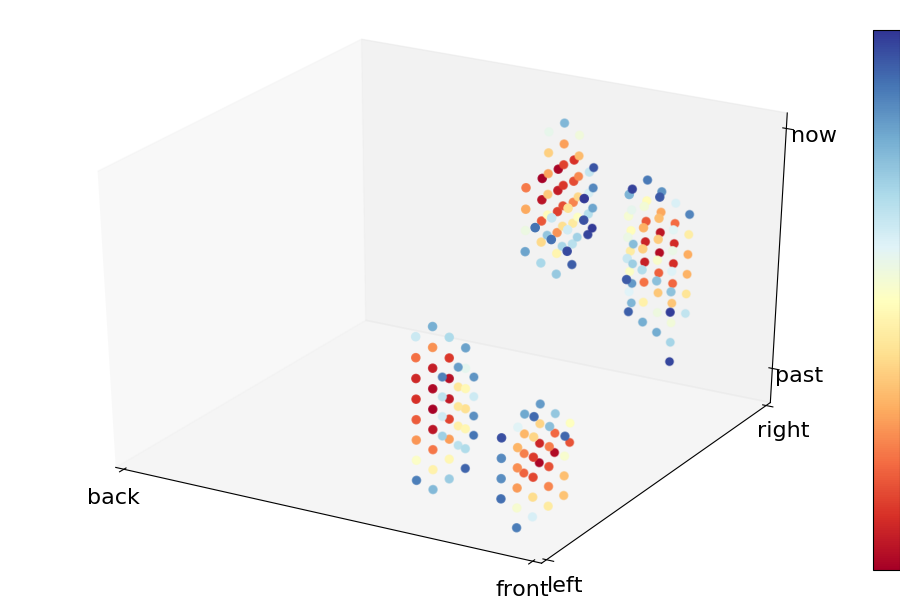} \label{ablation:a}}
        \subfigure[8.62]{\includegraphics[width=0.18\textwidth]{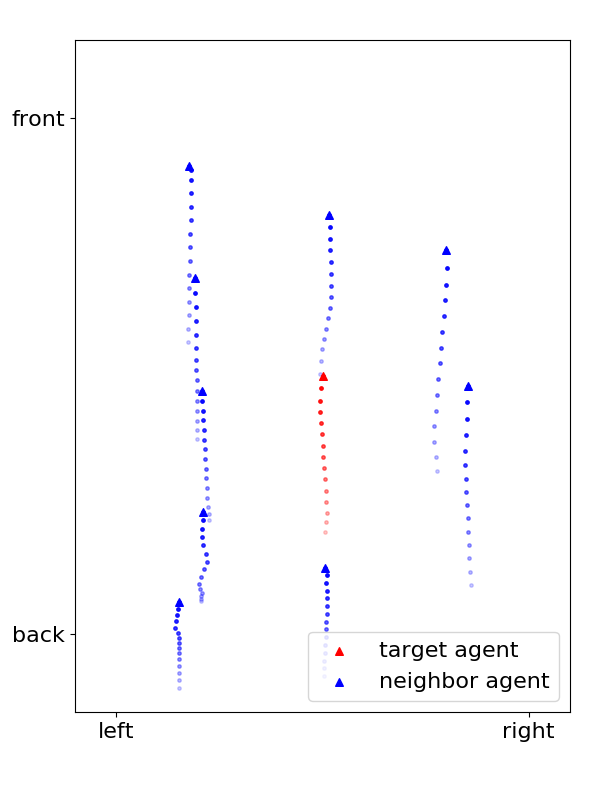} \label{ablation:b}}
        \subfigure[6.21]{\includegraphics[width=0.18\textwidth]{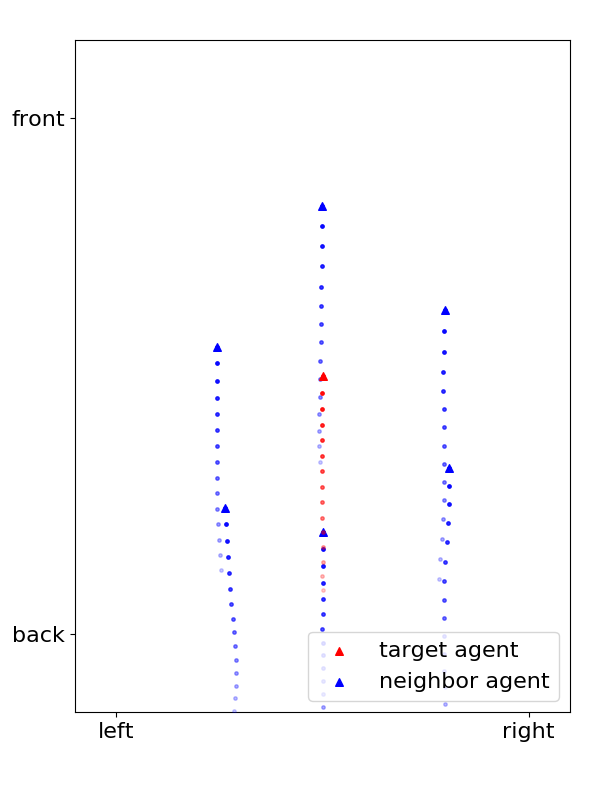} \label{ablation:c}}
        \subfigure[1.38]{\includegraphics[width=0.18\textwidth]{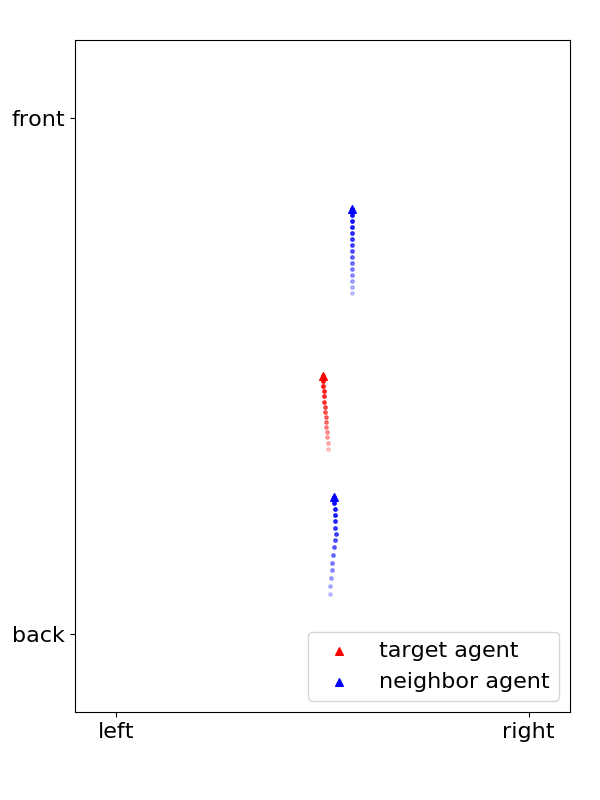} \label{ablation:d}}
        \subfigure[0.34]{\includegraphics[width=0.18\textwidth]{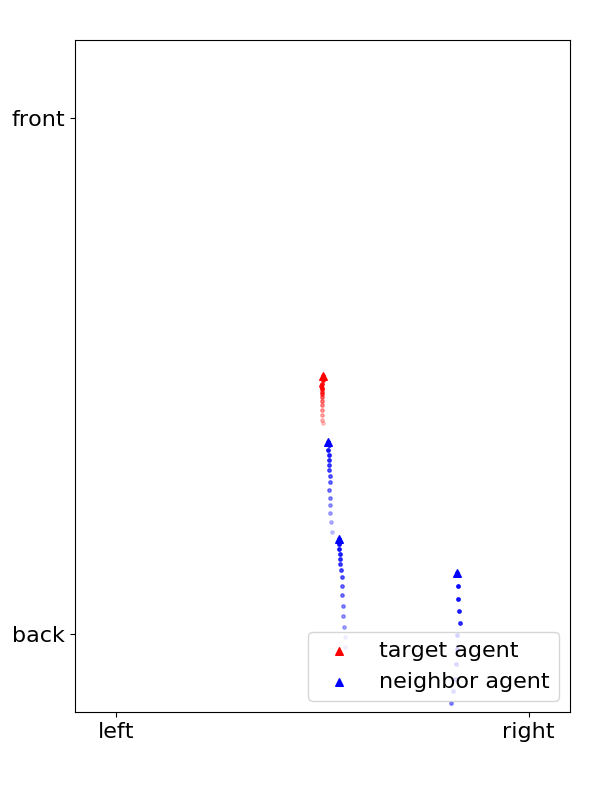} \label{ablation:e}}
    }
    \centerline{
        \subfigure[Partition for deceleration of front vehicle]{\includegraphics[width=0.3\textwidth]{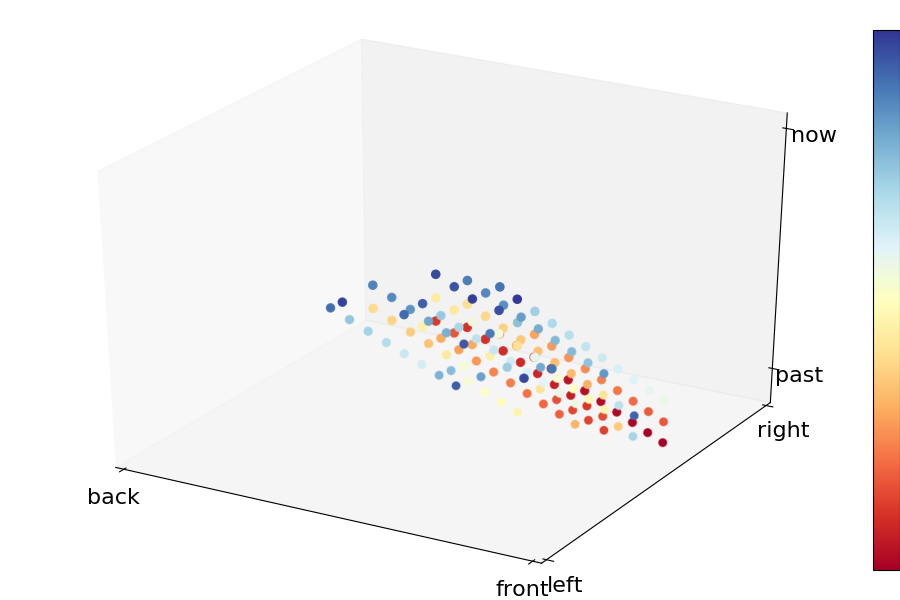} \label{ablation:f}}
        \subfigure[7.53]{\includegraphics[width=0.18\textwidth]{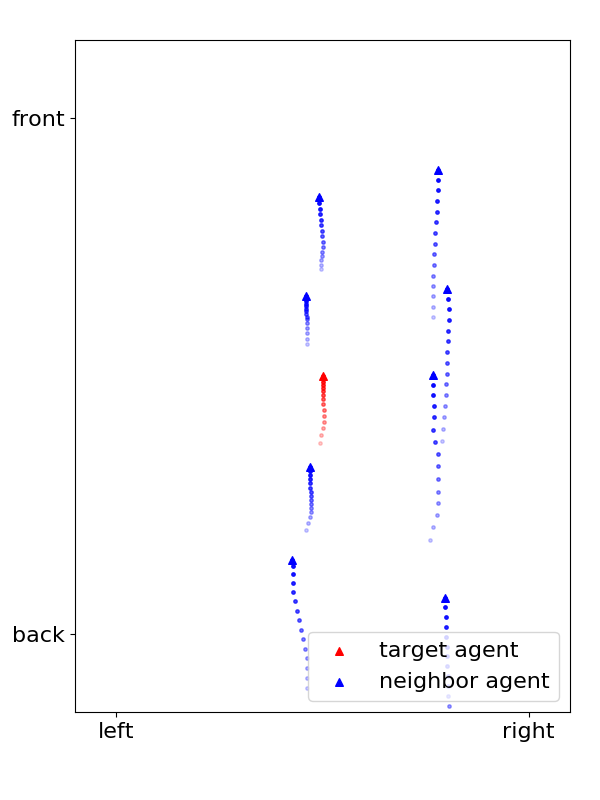} \label{ablation:g}}
        \subfigure[6.68]{\includegraphics[width=0.18\textwidth]{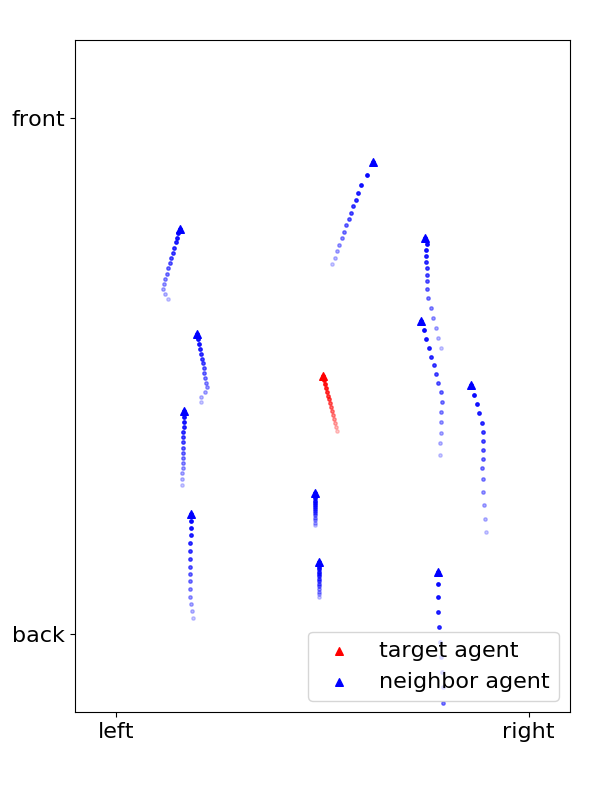} \label{ablation:i}}
        \subfigure[3.57]{\includegraphics[width=0.18\textwidth]{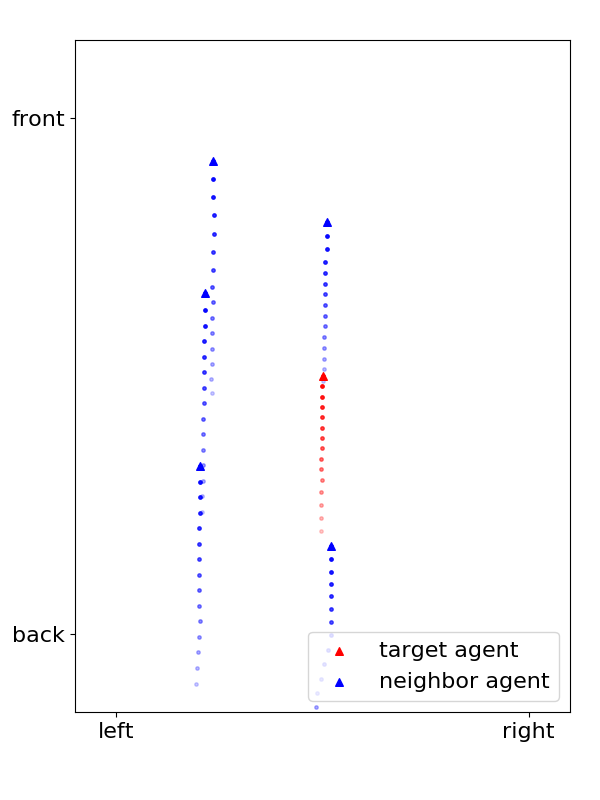} \label{ablation:h}}
        \subfigure[1.02]{\includegraphics[width=0.18\textwidth]{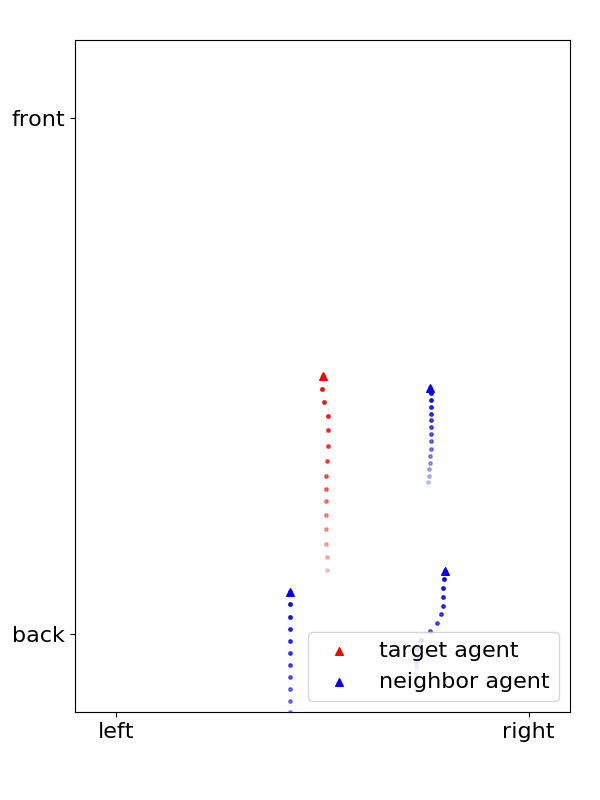} \label{ablation:j}}
    }
    \caption{
        (a) and (f) show activation patterns of two typical neurons in the pooled spatio-temporal features.
         The number in other subfigures indicates the value of activation of this neuron of the case.
    }\label{fig:partition}
\end{figure*}

Each dimension in the global feature actually corresponds to a partition in the feature space. Each of them represents one configuration of spatio-temporal context. And this partition is learnt end-to-end from the data. To intuitively understand these operation, we visualize several examples in Fig.~\ref{fig:partition} with NGSIM~\cite{colyar2007us} dataset.

We randomly choose two dimensions from the global context feature, and find the most influencing positions in this 3D space. 
We further pickup some typical cases and present their activation of this dimension in these figures.
For the case in the Fig.~\ref{ablation:a}, most points spread in the left front and right front area.
Observing the scenarios in Fig.~\ref{ablation:b} - Fig.~\ref{ablation:e},
we can find that the scenario with more agents appearing in left front and right front has larger value in the corresponding dimension of the global feature.
Fig.~\ref{ablation:f} - Fig.~\ref{ablation:j} demonstrates another case that the agent should focus on past status of the front vehicle. Examining the high activation of the cases, we can find it actually corresponds to the decelerating of front vehicle.
It is clearly seen that the learned partition of spatio-temporal space by our proposed method coincides with human interpretation.


\paragraph{Recursive Refinement}
Although this aggregator is capable to aggregate the overall environment around the target agent, the pooling operator is still a first-order aggregator.
It fails to model the second-order information such as interactions between agents.
As a simple solution, we concatenate the global context feature to every individual feature, and recursively apply the aforementioned steps.
In the second step, the embedding is aware of the status of individual agent and all the global context, thus could capture the interactions.

We summarize the pseudocodes of our encoder structure in Alg.~\ref{alg:code}. It can be easily implemented with several lines of code. 
\begin{algorithm}[!htb]
    \caption{Pseudocode of Spatio-temporal pooling in a PyTorch-like style.}
    \label{alg:code}
    \algcomment{\fontsize{7.2pt}{0em}\selectfont \texttt{pooling}: max pooling; \texttt{cat}: concatenation.
    }
    \definecolor{codeblue}{rgb}{0.25,0.5,0.5}
    \lstset{
      backgroundcolor=\color{white},
      basicstyle=\fontsize{7.2pt}{7.2pt}\ttfamily\selectfont,
      columns=fullflexible,
      breaklines=true,
      captionpos=b,
      commentstyle=\fontsize{7.2pt}{7.2pt}\color{codeblue},
      keywordstyle=\fontsize{7.2pt}{7.2pt},
    }

    \begin{lstlisting}[language=python]

        def embedding(x, n_layers):
            # x: [N, K, Q] => [N, K, shape of weight]
            for layer in range(n_layers):
                x = Linear(x)
                x = BatchNorm(x)
                x = ReLU(x)
            return x

        def STPooling(x):
            '''spatial temporal context extractor
            Args:
                x: [N, K, Q]. 
                   N samples with K nbrs,
                   snapshot's dimension Q, 
            Return:
                context: [N, K]
            '''
            e1 = embedding(x, n_layers=2)
            context1 = pooling(e1, dim=2)

            context2 = context1.repeat(1, x.size[-1])
            context2 = cat((context2, x), dim=2)
            e2 = embedding(context2, n_layers=2)
            context2 = pooling(e2, dim=2).squeeze()
            return context2
    \end{lstlisting}
    \end{algorithm}

\subsection{Future Trajectory Decoder}\label{decoder}
The decoder should be designed to satisfy various demands from downstream modules like planning.
We believe that when the spatio-temporal context is well modeled, off-the-shelf decoder can still achieve superior results. 
As a default option, we feed the encoded spatio-temporal feature into a standard LSTM. At each time step, we minimize the $l_2$ loss between predicted 2D positions and the ground-truth.
If desired, it is easy to extend to a stochastic decoder. We could simply inject a zero-mean Gaussian noise to the encoded feature $C$ in each decoding process, and using variety loss as  supervision as in\cite{gupta2018social, zhao2019multi, huang2019stgat}.

In section \ref{sec:experiment}, extensive experiments show that our method can achieve state-of-the-art performance in both deterministic and stochastic prediction tasks despite the simplicity of our decoder structure.

\subsection{Discussion}
In this section, we highlight several key advantages of our proposed method, and compare with other state-of-the-art method.
\paragraph{RNN free}
One benefit is that we do not rely on a fixed-length history trajectory, thus we can better handle the missing data due to the false negative of detection or tracker interruption. In contrast, current RNN based methods~\cite{gupta2018social, deo2018convolutional} have to eliminate the incomplete history of an agent because RNN can not differentiate the exact time step of the remaining trajectory, which reduces the amount of effective context one can utilize when modeling the behavior of the target agent.
Second, RNN-based encoder for varying length trajectories is hard to be batch parallelled.
Padding dummy values can be a solution, but it will introduce noise and affect performance.
Besides, RNN-based encoder does not work when the time gap of the history sequence is not fixed, which is usually the case in real world autonomous driving scenarios because of the latency of the upstream fusion and perception modules.
On the contrary, the time dimension in our input representation can be a dynamic floating point, which is much more flexible.


\paragraph{Comparisons with other social pooling methods}
Although there are several works that also use pooling module to model social interaction~\cite{alahi2016social}, their methods and motivation are different from us.
Their method needs to partition the space manually, and compute additional occupancy grids for each agent, while ours has already embedded and partitioned both space and time adaptively thanks to the unified spatio-temporal representation. Therefore, a simple global pooling is enough. 

\paragraph{The importance of modeling cross time step social interaction}
People need the reaction time to handle the sudden circumstance on the road. 
For example, there exists a delay before one can step on brakes when the front car slows down.
In other words, an agent's behavior may be affected by the actions of other agents in previous time steps.
Thus, extracting the cross-time step interactions among agents becomes necessary.


%% file: content/experiment.tex
\section{Experiment}\label{sec:experiment}
We compare our proposed method with other state-of-the-art methods in two recently published trajectory datasets: ApolloScape \cite{huang2018apolloscape, ma2019trafficpredict} and Argoverse \cite{chang2019argoverse}.
Both of these datasets are collected from the first-person perspective by sensor-equipped acquisition cars in real-world driving scenarios, but they focus on different problems in the trajectory prediction field.
We use the official metrics for these datasets and compare our proposed \pmethodb~ with the state-of-the-arts.
In addition, we construct variants of our method for ablation study and qualitative analysis to show how \pmethodb~ can model complex spatio-temporal context in the high dimensional space.


\subsection{Dataset and Experiment Setting}
In this section, we detail the datasets as well as the corresponding experiment settings.

\begin{table*}[!bt]
    \begin{tabular}{@{}
        +m{1.0in}<{\centering}
        Ym{0.6in}<{\centering}Ym{0.6in}<{\centering}
        Ym{0.6in}<{\centering}Ym{0.6in}<{\centering}|Ym{0.6in}<{\centering}
        Ym{0.5in}<{\centering}Ym{0.5in}<{\centering}Ym{0.5in}<{\centering}
        @{}}
        \toprule
        & \multicolumn{3}{c} {(a) ADE (3s)}  & \multicolumn{5}{c}{(b) FDE (3s) }  \\
         Methods & vehicles & pedestrians & bicycles & weighted & vehicles & pedestrians & bicycles & weighted \\
        \midrule{}
        StarNet\cite{zhu2019starnet}                    &      2.38      &      0.78      &      1.86      &   1.34   &      4.28       &      1.51       &       3.46  &   2.49      \\
        TrafficPredict\cite{ma2019trafficpredict}                         &      7.94      &      7.18      &      12.88      &      8.58 &      12.77       &      11.12       &      22.79   &   24.22       \\
        LSTM  &      2.88      &      0.94      &      2.09      &    1.58   &      5.25       &      1.84       &      3.87    &  2.97       \\
        CV  &      2.59      &      0.81      &      2.17      &      1.47   &      4.64       &      1.58       &      4.02      &  2.73  \\
        \pmethodb                            &      \textbf{2.10}      &      \textbf{0.75}      &      \textbf{1.77}      &      \textbf{1.24}     &      \textbf{3.65}      &       \textbf{1.44}        &       \textbf{3.14}    &     \textbf{2.25}       \\
        \bottomrule
    \end{tabular}
    \caption{Performance on ApolloScape trajectory dataset. All results are recorded from ApolloScape public leaderboard. Our method rank first in all public methods currently. (a) shows the average distance error (ADE) of different algorithms on diverse traffic-agents. (b) shows the final distance error (FDE) in 3 second.}\label{tab:apollo}
\end{table*}

\paragraph{ApolloScape~\cite{huang2018apolloscape}}
ApolloScape dataset is recorded in urban streets by various sensors (e.g. LiDAR, radar, camera).
This dataset provides 3 seconds as trajectory history and aims to predict the next 3 seconds at 0.5s interval.
The scene consists of heterogeneous traffic agents, including pedestrians, vehicles and bicycles.
So the spatio-temporal context is much more complex to model.
To give a fair comparison for this heterogeneous traffic-agent dataset, we follow the same experiment setting in ApolloScape challenge \footnote{ApolloScape Challange \url{http://apolloscape.auto/leader_board.html}}.
We submit our result to the challenge leaderboard, and compare it with other submitted methods.
To measure the performance of algorithms, we report the prediction error of each type of traffic-agents.
The main metrics used for this dataset are Average Distance Error (ADE) and Final Distance Error (FDE).
\begin{itemize}
    \item ADE: mean Euclidean distance between predicted coordinates and the ground truth over all time steps.
    $$\mathrm{ADE}=\frac{\sum_{n=1}^{N} \sum_{t=1}^{T}\left\|\tilde{x}_{n}^{t}-x_{n}^{t}\right\|_{2}}{N * T}$$
    \item FDE: Euclidean distance between the predicted coordinates and the ground truth at the final prediction timestep.
    $$\mathrm{FDE}=\frac{\sum_{n=1}^{N} \left\|\tilde{x}_{n}^{T}-x_{n}^{T}\right\|_{2}}{N}$$
\end{itemize}

Based on these metrics, we report a weighted sum of ADE (WSADE) and weighted sum of FDE (WSFDE) by assigning coefficient $0.20, 0.58, 0.22$ to pedestrians, vehicles, bicycles respectively as in the ApolloScape Challenge.
\begin{equation*}
\begin{aligned}
    \mbox{WSADE} &= 0.2 \cdot \mbox{ADE}_v + 0.58 \cdot \mbox{ADE}_p + 0.22 \cdot \mbox{ADE}_b\\
    \mbox{WSFDE} &= 0.2 \cdot \mbox{FDE}_v + 0.58 \cdot \mbox{FDE}_p + 0.22 \cdot \mbox{FDE}_b
\end{aligned}
\end{equation*}
WSADE is also the metric by which the ApolloScape challenge is ranked.
Because the difference of intrinsic behavior between pedestrians and vehicles, we separate the model for pedestrians by increasing the pedestrians' weight in the loss function.

\paragraph{Argoverse~\cite{chang2019argoverse}}
Argoverse is a large-scale autonomous driving dataset, containing 320 hours of data as well as rich map information.
The given vector map is a semantic graph that provides detailed lane information that an agent might follow.
Therefore we can get the multi-modal future trajectories explicitly with the help of multiple candidate lane centerlines.
The collected trajectories in this dataset are individual 5 seconds trajectory segments. The first 2 seconds are used as history and predict spatial locations of the vehicles for up to 5 seconds.
The traffic-agent type of Argoverse is not provided, so it is not included in our input representation.

We follow the official metric to benchmark multiple predictions on this dataset.
The metric we choose is Minimum over N (MoN) metric as in previous works \cite{gupta2018social, lee2017desire}. It computes the error between the ground truth and the closet trajectory provided in the N output predictions.
Specifically, we evaluate the top-N ADE and FDE with $N=6$.
To generate multi-modal future trajectory, we use the basic implementation offered by Argoverse baseline\footnote{Argoverse baseline implementation: \url{https://github.com/jagjeet-singh/argoverse-forecasting}}.
In training, we choose a \textit{2-d curvilinear coordinate system} with axes tangential and perpendicular to the most possible centerline of the trajectory.
At inference, we generate diverse future trajectories by using a different centerline as the 2-d curvilinear coordinate system. With multiple candidate centerlines, we can define various origins and reference frames to predict diverse futures.
\paragraph{NGSIM}
Besides real-world autonomous driving dataset, we also evaluate our method on a widely used trajectory dataset Next Generation Simulation (NGSIM) dataset \cite{colyar2007us}.
It consists of 45 minutes of highway driving trajectories at 10Hz for each roadway and contains various traffic conditions and diverse interactions among different traffic-agents.
Note that trajectories in this dataset are recorded from fixed bird-eye view cameras.
It can observe traffic agents without any coverage or missing data, which is inconsistent with the real autonomous driving scenarios. Thus we only list the results for reference.
We adopt the same experiment setting as \cite{deo2018convolutional} to use $3s$ as history and predict $5s$ for future trajectory.
The evaluation metric is root mean square error (RMSE) in meters over all future timesteps.

\subsection{Implementation Details}\label{sec:implementation_details}
We set the filter number of the fully connected layer to
128. Also the dimension of the hidden state for decoder
LSTM keeps the same too. We iteratively train the network
with a batch size of 128 for 50 epochs using Adam with an
initial learning rate of 0.0003. We set the weight decay to 0.0001.
The basic input of our model is 2D location, velocity, discrete time, and ID.
We also add the type of agents on the ApolloScape dataset since in urban scenarios usually mixed types of agents are presented.


%% file: content/result.tex
\subsection{Results}

We compare our proposed method with several state-of-the-art algorithms and baseline models in this section.

\begin{figure*}[!htb]
        \centerline{
        \subfigure{\includegraphics[width=0.25\textwidth]{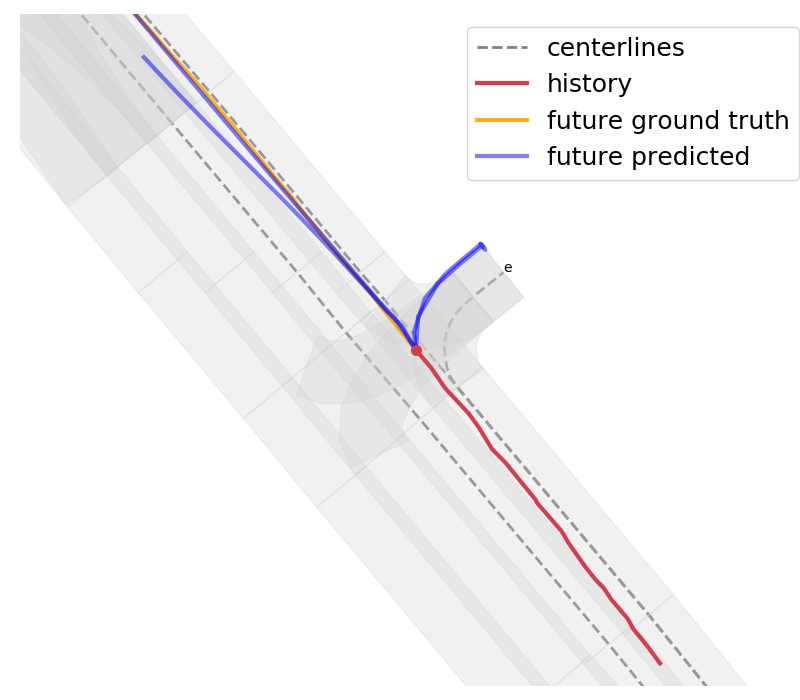} \label{dd}}
        \subfigure{\includegraphics[width=0.25\textwidth]{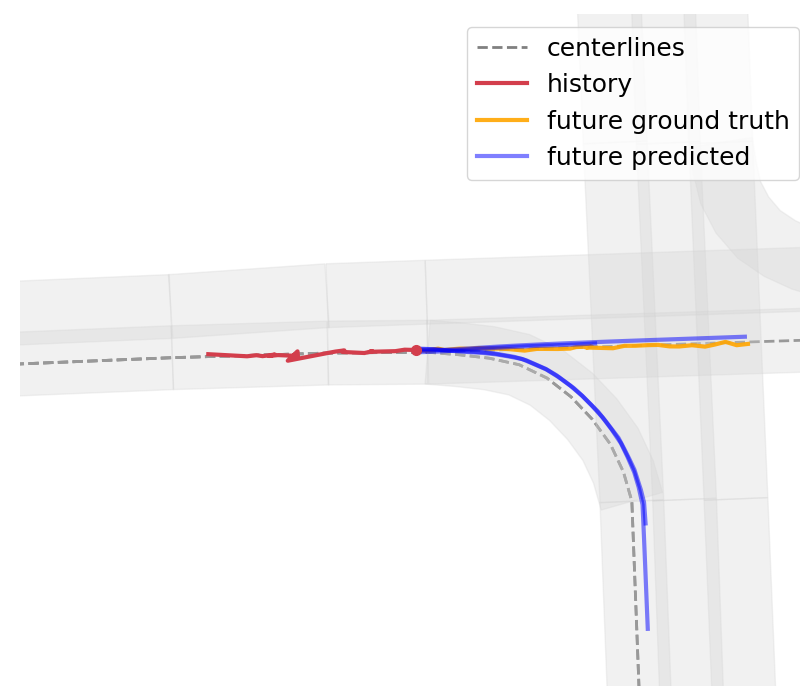} \label{dd}}
        \subfigure{\includegraphics[width=0.25\textwidth]{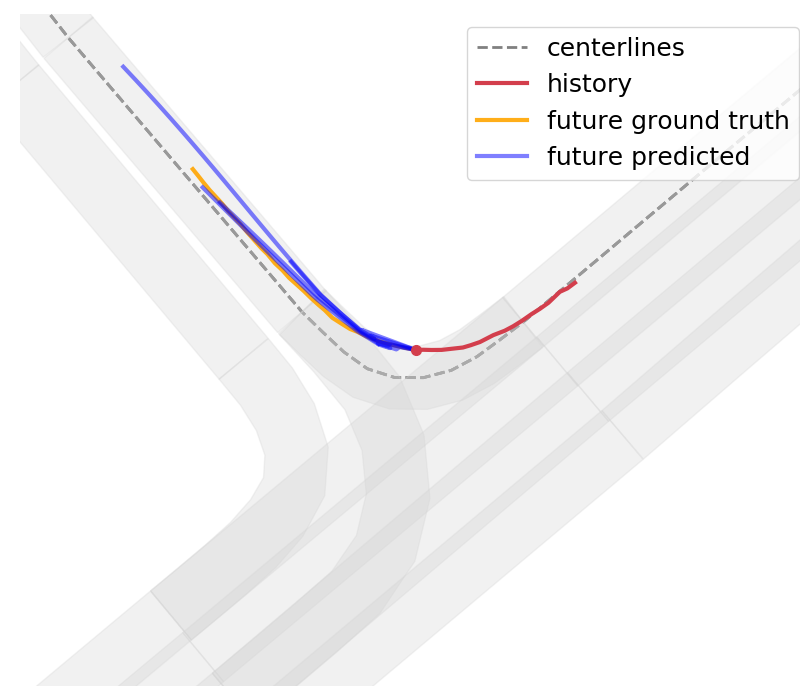} \label{ss}}
        \subfigure{\includegraphics[width=0.25\textwidth]{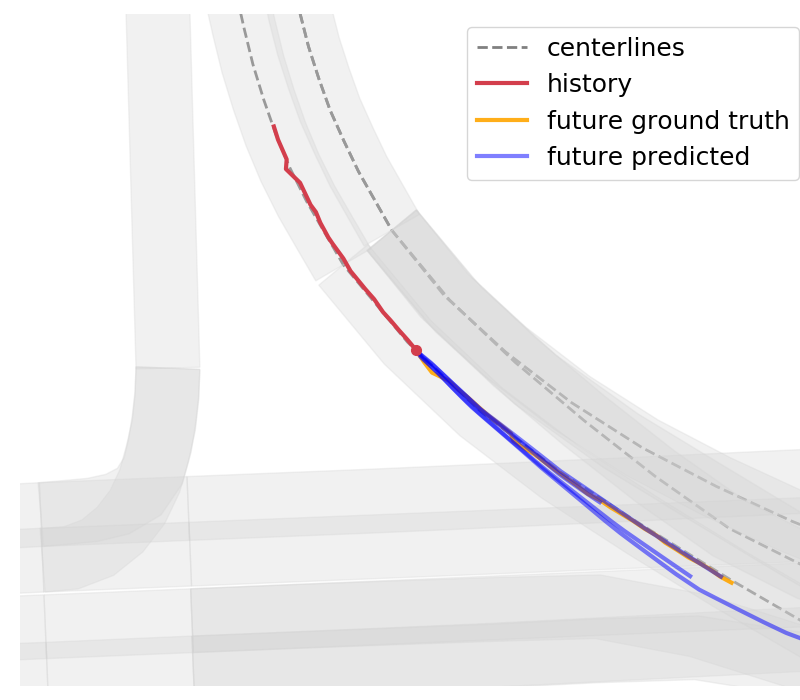} \label{ss}}
        }
        \caption{
            Visualization of diverse trajectory predictions on Argoverse dataset. With basic implementation, we can still generate multiple diverse trajectory using the well modeled spatio-temporal context.
        }\label{fig:multiple_argo}
\end{figure*}

\paragraph{ApolloScape}
State-of-the-art algorithms on ApolloScape and baseline models are shown as follows.
The results are excerpted from ApolloScape leaderboard to make sure a fair comparison.
\begin{itemize}
    \setlength\itemsep{0.1em}
    \item StarNet\cite{zhu2019starnet}: 
    The champion of trajectory prediction challenge in ``CVPR 2019 Workshop on Autonomous Driving — Beyond Single Frame Perception'', which utilizes a centralized hub network to model spatio-temporal contexts with low time complexity. 
    \item TrafficPredict\cite{ma2019trafficpredict}: An LSTM-based algorithm which focuses on predicting trajectories for heterogeneous traffic-agents in urban environment.
    \item LSTM: A simple LSTM encoder-decoder which only models each agent's temporal context independently.
    \item CV(constant velocity): A baseline which uses a constant velocity Kalman filter to predict trajectories in the future.
\end{itemize}
In Table \ref{tab:apollo}, the results show that our proposed method achieves best performance in every metric and every agent type.
Especially, compared with the current state-of-the-art method StarNet which is an LSTM-based attention network, we observe a \textbf{0.63m} improvement in FDE for the vehicle prediction task.
Although they have leveraged a hub network to model agents' interactions, the encoded context is still worse than ours.
Our method improves considerably compared with all other methods of all three kinds of traffic agents.

\paragraph{Argoverse}
\begin{table}[!htb]
    \centering
    \begin{tabular*}{0.45\textwidth}{
                >{}Ym{1.3in}
                >{}Ym{0.6in}<{\centering}  Ym{0.6in}<{\centering}
        }
        \toprule
        \rowstyle{}
        Methods &\multicolumn{1}{c}{{$3s$ ADE}}  &\multicolumn{1}{c}{{$3s$ FDE}}  \\
        \midrule{}                                                                                                                 
        NN + map  &  2.28  & 4.80  \\ 
        LSTM + map   &  2.25  &  4.67  \\
        LSTM + social + map   &  2.46  & 4.67  \\ 
        \pmethodb~  &  \textbf{1.47} &  \textbf{2.94}   \\ 
        \bottomrule
    \end{tabular*}
    \caption{Evaluations of \pmethodb~and multiple baselines on the Argoverse dataset. Error is minimum over six samples.}
    \label{tab:argo}
\end{table}
We compare our method with several baselines from Argoverse.
\begin{itemize}
    \item Nearest Neighbor with map (NN + map): Weighted Nearest Neighbor regression trajectories where trajectories are queried by coordinate in the curvilinear coordinate system.
    \item LSTM + map: standard LSTM encoder-decoder structure, which uses map information to define multiple curvilinear coordinate systems. Therefore, it can generate multiple possible future trajectories.
    \item LSTM + social + map : Similar to LSTM + map, but with additional hand-crafted social features. Social features include minimum distance to the vehicle in front and in back, and number of neighbors.
\end{itemize}

Table~\ref{tab:argo} shows that \pmethodb~outperforms all baselines on both $3s$ ADE and $3s$ FDE metrics.
We also provide qualitative results to visualize how \pmethodb~can predict multiple possible future trajectories according to maps (as shown in Fig. \ref{fig:multiple_argo}). Once the spatio-temporal context is well modeled, we can easily generate multiple multi-modal trajectories with additional map information under different complex conditions.


\paragraph{NGSIM}
We report the performance of \pmethodb~and several state-of-the-art algorithms on the NGSIM dataset (see Table. \ref{tab:ngsim}).
Only deterministic version result is reported.
In addition, we construct a variant of our method denoted as \textit{\pmethodb-180} which doubles the context range by enlarging the longitudinal direction range from $\pm 90$ to $\pm 180$.

Without any doubt, \pmethodb~outperforms any other algorithm in the same setting. In particular, the results of $\pm 180$ improves dramatically. It outperforms the standard version, especially in the result of $5s$.
We believe that it is owe to the richer spatio-temporal context provided, and our unified modeling method could digest these information seamlessly.

\begin{table*}[!htb]
	\centering
		\begin{tabular}{@{}
			+m{0.6in}<{\centering}
			Ym{0.4in}<{\centering}Ym{0.4in}<{\centering}
			Ym{0.4in}<{\centering}Ym{0.4in}<{\centering}
			Ym{0.4in}<{\centering}Ym{0.4in}<{\centering}
            Ym{0.4in}<{\centering}Ym{0.4in}<{\centering}
			Ym{0.4in}<{\centering}Ym{0.4in}<{\centering}Ym{0.4in}<{\centering}
			Ym{0.4in}<{\centering}Ym{0.6in}<{\centering}Ym{0.4in}<{\centering}Ym{0.4in}<{\centering}
			@{}}
			\toprule
            \noalign{\smallskip}
            \rowstyle{\small} 
            Time   & CV & CV-GMM\cite{deo2018would} & GAIL-GRU\cite{kuefler2017imitating}   & LSTM   & MATF~\cite{zhao2019multi}  & CS-LSTM~\cite{deo2018convolutional}  &  S-LSTM~\cite{alahi2016social}  &\pmethodb~  & \pmethodb-180  \\
			\midrule{}
			1$s$  &  0.73  &  0.66 & 0.69 & 0.68 & 0.67   & 0.61 & 0.65 & \textbf{0.58} & 0.56     \\ 
			2$s$  &  1.78  &  1.56 & 1.56 & 1.65 & 1.51   & 1.27 & 1.31 & \textbf{1.20} & 1.15     \\ 
			3$s$  &  3.13  &  2.75 & 2.75 & 2.91 & 2.51   & 2.09 & 2.16 & \textbf{1.96} & 1.82     \\ 
			4$s$  &  4.78  &  4.24 & 4.24 & 4.46 & 3.71   & 3.10 & 3.25 & \textbf{2.92} & 2.58     \\ 
			5$s$  &  6.68  &  5.99 & 5.99 & 6.27 & 5.12   & 4.37 & 4.55 & \textbf{4.12} & 3.45     \\ 
			\bottomrule
        \end{tabular}
        \caption{
     Comparison with different methods. Root mean square errors (RMSE) in meters from $1s$ to $5s$ are reported.
	}\label{tab:ngsim}
\end{table*}

\subsection{Ablation Study}
In this section, we design several variants of \pmethodb~to investigate how different setting influences the final performance.
All methods are evaluated on Argoverse validation set. The metric in ablation study is top-N ADE and FDE with N=1 to control all possible factors except the encoder. 
\paragraph{Input Representation} We study the feature used in the input representation of \pmethodb.
Table.~\ref{tab:ablation} summeraizes the results.
Note that the velocity comes from the differential of the provided position.
As can be noticed, velocity only brings in marginal improvement. 
We owe this observation to that \pmethodb~ can model the relationship between different agents across different time steps, thus speed feature has already been implicitly learned. This property is especially useful when the observations from upstream are noisy, in which the results of explicit differential are not reliable.
  
\begin{table}[!tb]
	\centering
		\begin{tabular}{@{}
			Ym{0.4in}<{\centering}Ym{0.4in}<{\centering}
			Ym{0.4in}<{\centering}Ym{0.4in}<{\centering}
			Ym{0.4in}<{\centering}Ym{0.4in}<{\centering}
			@{}}
			\toprule
            \noalign{\smallskip}
            \rowstyle{\small} 
            Position &  Time &  Velocity   &   3s ADE  & 3s FDE \\ 
			\midrule{}
			\checkmark  &             &             &  3.46  &  7.60  \\ 
			\checkmark  &  \checkmark &             &  2.71  &  6.01 \\ 
			\checkmark  &  \checkmark &  \checkmark &  2.68  &  5.97 \\ 
			\bottomrule
        \end{tabular}
        \caption{
	Results of ablation studies on the Argoverse validation set.
	}\label{tab:ablation}
\end{table}

\begin{figure}[!tb]
	\centering
	\includegraphics[width=0.44\textwidth]{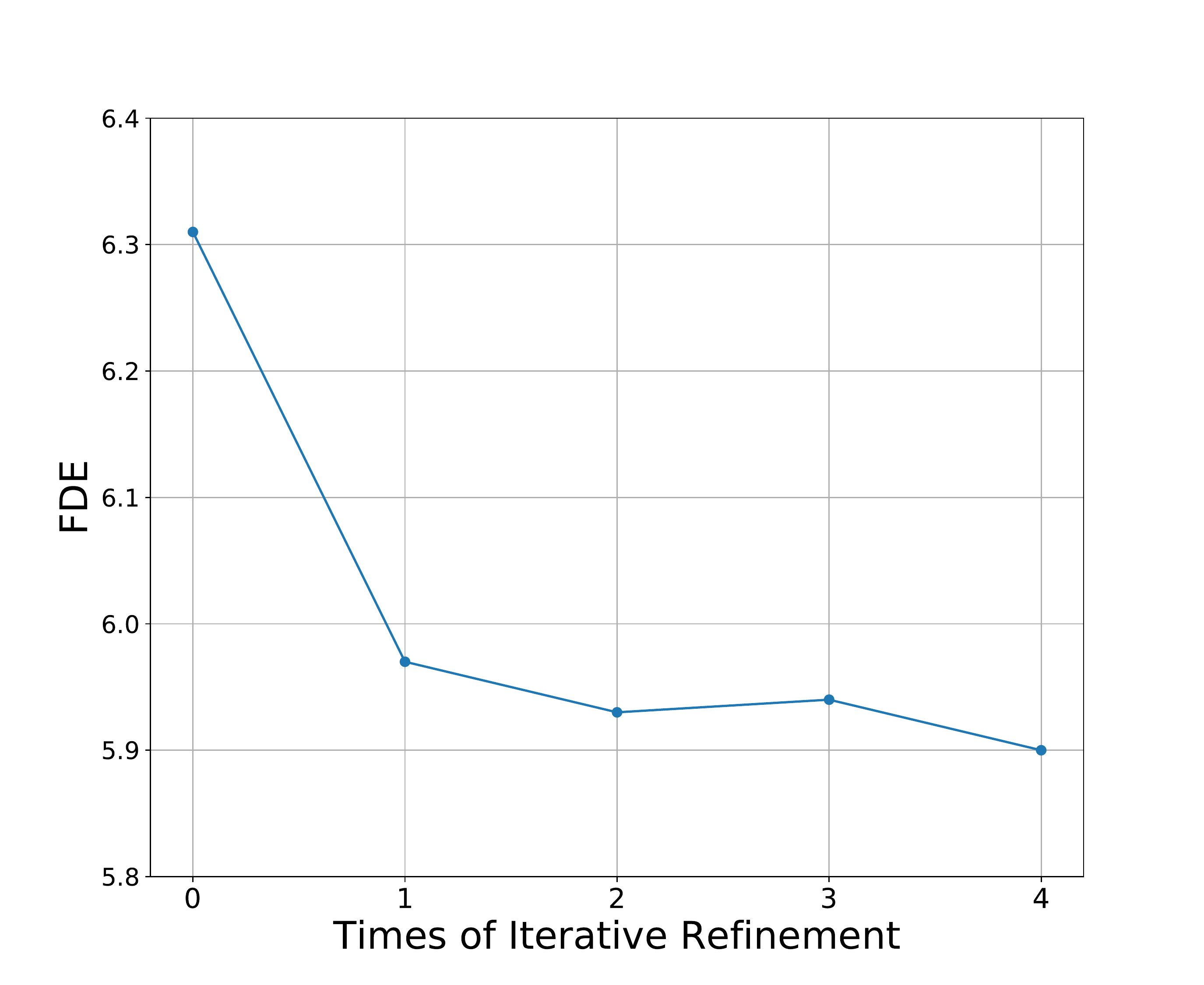}
	\caption[width=\columnwidth]{
		Results on the Argoverse validation set under the different time of iterative refinement.
   }\label{fig:number_refine_blocks}
\end{figure}

\paragraph{Number of Times of Iterative Refinement} We also conduct experiments to investigate the time of recursive refinement. 
The results in Fig.~\ref{fig:number_refine_blocks} show that when the time of refinement exceeds two, the performance becomes saturated. This may suggest that the gain of modeling even higher-order interactions more than two is marginal.

%% file: content/conclusion.tex
\section{Conclusion}
In this paper, we have presented \pmethodb~for trajectory prediction problem, which integrates 2D locations and discrete time space into one unified 3D space, then learn the spatio-temporal context end-to-end.
Despite the simplicity, it still shows state-of-the-art performance on various trajectory prediction datasets.
We hope our method could be a strong baseline to trajectory prediction field and these encouraging results could inspire more advanced methods on the spatio-temporal context representation.

%% file: main2.bbl
\begin{thebibliography}{10}
\providecommand{\url}[1]{#1}
\csname url@rmstyle\endcsname
\providecommand{\newblock}{\relax}
\providecommand{\bibinfo}[2]{#2}
\providecommand\BIBentrySTDinterwordspacing{\spaceskip=0pt\relax}
\providecommand\BIBentryALTinterwordstretchfactor{4}
\providecommand\BIBentryALTinterwordspacing{\spaceskip=\fontdimen2\font plus
\BIBentryALTinterwordstretchfactor\fontdimen3\font minus
  \fontdimen4\font\relax}
\providecommand\BIBforeignlanguage[2]{{%
\expandafter\ifx\csname l@#1\endcsname\relax
\typeout{** WARNING: IEEEtran.bst: No hyphenation pattern has been}%
\typeout{** loaded for the language `#1'. Using the pattern for}%
\typeout{** the default language instead.}%
\else
\language=\csname l@#1\endcsname
\fi
#2}}

\bibitem{gupta2018social}
A.~Gupta, J.~Johnson, L.~Fei-Fei, S.~Savarese, and A.~Alahi, ``Social {GAN}:
  {S}ocially acceptable trajectories with generative adversarial networks,'' in
  \emph{CVPR}, 2018, pp. 2255--2264.

\bibitem{deo2018convolutional}
N.~Deo and M.~M. Trivedi, ``Convolutional social pooling for vehicle trajectory
  prediction,'' in \emph{CVPRW}, 2018, pp. 1468--1476.

\bibitem{zhang2019sr}
P.~Zhang, W.~Ouyang, P.~Zhang, J.~Xue, and N.~Zheng, ``{SR-LSTM}: {State}
  refinement for {LSTM} towards pedestrian trajectory prediction,'' in
  \emph{CVPR}, 2019, pp. 12\,085--12\,094.

\bibitem{li2019grip}
X.~Li, X.~Ying, and M.~C. Chuah, ``{GRIP}: {G}raph-based interaction-aware
  trajectory prediction,'' \emph{arXiv preprint arXiv:1907.07792}, 2019.

\bibitem{alahi2016social}
A.~Alahi, K.~Goel, V.~Ramanathan, A.~Robicquet, L.~Fei-Fei, and S.~Savarese,
  ``Social {LSTM}: {H}uman trajectory prediction in crowded spaces,'' in
  \emph{CVPR}, 2016, pp. 961--971.

\bibitem{vemula2018social}
A.~Vemula, K.~Muelling, and J.~Oh, ``Social {attention}: {Modeling} attention
  in human crowds,'' in \emph{ICRA}, 2018, pp. 1--7.

\bibitem{zhao2019multi}
T.~Zhao, Y.~Xu, M.~Monfort, W.~Choi, C.~Baker, Y.~Zhao, Y.~Wang, and Y.~N. Wu,
  ``Multi-agent tensor fusion for contextual trajectory prediction,'' in
  \emph{CVPR}, 2019, pp. 12\,126--12\,134.

\bibitem{messaoud2019non}
K.~Messaoud, I.~Yahiaoui, A.~Verroust-Blondet, and F.~Nashashibi, ``Non-local
  social pooling for vehicle trajectory prediction,'' 2019.

\bibitem{kosaraju2019social}
V.~Kosaraju, A.~Sadeghian, R.~Mart{\'\i}n-Mart{\'\i}n, I.~Reid, S.~H.
  Rezatofighi, and S.~Savarese, ``Social-{B}i{GAT}: {M}ultimodal trajectory
  forecasting using bicycle-gan and graph attention networks,'' \emph{arXiv
  preprint arXiv:1907.03395}, 2019.

\bibitem{huang2019stgat}
Y.~Huang, H.~Bi, Z.~Li, T.~Mao, and Z.~Wang, ``{STGAT}: {M}odeling
  spatial-temporal interactions for human trajectory prediction,'' in
  \emph{ICCV}, 2019, pp. 6272--6281.

\bibitem{qi2017pointnet}
C.~R. Qi, H.~Su, K.~Mo, and L.~J. Guibas, ``Pointnet: {D}eep learning on point
  sets for 3{D} classification and segmentation,'' in \emph{CVPR}, 2017, pp.
  652--660.

\bibitem{helbing1995social}
D.~Helbing and P.~Molnar, ``Social force model for pedestrian dynamics,''
  \emph{Physical {R}eview E}, vol.~51, no.~5, p. 4282, 1995.

\bibitem{luber2010people}
M.~Luber, J.~A. Stork, G.~D. Tipaldi, and K.~O. Arras, ``People tracking with
  human motion predictions from social forces,'' in \emph{ICRA}, 2010, pp.
  464--469.

\bibitem{ferrer2013robot}
G.~Ferrer, A.~Garrell, and A.~Sanfeliu, ``Robot companion: {A} social-force
  based approach with human awareness-navigation in crowded environments,'' in
  \emph{IROS}, 2013, pp. 1688--1694.

\bibitem{yamaguchi2011you}
K.~Yamaguchi, A.~C. Berg, L.~E. Ortiz, and T.~L. Berg, ``Who are you with and
  where are you going?'' in \emph{CVPR}, 2011, pp. 1345--1352.

\bibitem{choi2013understanding}
W.~Choi and S.~Savarese, ``Understanding collective activities of people from
  videos,'' \emph{IEEE {T}ransactions on {P}attern {A}nalysis and {M}achine
  {I}ntelligence}, vol.~36, no.~6, pp. 1242--1257, 2013.

\bibitem{ding2019predicting}
W.~Ding, J.~Chen, and S.~Shen, ``Predicting vehicle behaviors over an extended
  horizon using behavior interaction network,'' \emph{arXiv preprint
  arXiv:1903.00848}, 2019.

\bibitem{ma2019trafficpredict}
Y.~Ma, X.~Zhu, S.~Zhang, R.~Yang, W.~Wang, and D.~Manocha, ``Trafficpredict:
  {Trajectory} prediction for heterogeneous traffic-agents,'' in \emph{AAAI},
  2019, pp. 6120--6127.

\bibitem{kim2017probabilistic}
B.~Kim, C.~M. Kang, J.~Kim, S.~H. Lee, C.~C. Chung, and J.~W. Choi,
  ``Probabilistic vehicle trajectory prediction over occupancy grid map via
  recurrent neural network,'' in \emph{ITSC}, 2017, pp. 399--404.

\bibitem{yi2016pedestrian}
S.~Yi, H.~Li, and X.~Wang, ``Pedestrian behavior understanding and prediction
  with deep neural networks,'' in \emph{CVPR}, 2016, pp. 263--279.

\bibitem{li2019way}
Y.~Li, ``Which way are you going? {I}mitative decision learning for path
  forecasting in dynamic scenes,'' in \emph{CVPR}, 2019, pp. 294--303.

\bibitem{cui2019multimodal}
H.~Cui, V.~Radosavljevic, F.-C. Chou, T.-H. Lin, T.~Nguyen, T.-K. Huang,
  J.~Schneider, and N.~Djuric, ``Multimodal trajectory predictions for
  autonomous driving using deep convolutional networks,'' in \emph{ICRA}, 2019,
  pp. 2090--2096.

\bibitem{djuric2018motion}
N.~Djuric, V.~Radosavljevic, H.~Cui, T.~Nguyen, F.-C. Chou, T.-H. Lin, and
  J.~Schneider, ``Motion prediction of traffic actors for autonomous driving
  using deep convolutional networks,'' \emph{arXiv preprint arXiv:1808.05819},
  2018.

\bibitem{nikhil2018convolutional}
N.~Nikhil and B.~Tran~Morris, ``Convolutional neural network for trajectory
  prediction,'' in \emph{ECCV}, 2018.

\bibitem{choi2019looking}
C.~Choi and B.~Dariush, ``Looking to relations for future trajectory
  forecast,'' \emph{arXiv preprint arXiv:1905.08855}, 2019.

\bibitem{chai2019multipath}
Y.~Chai, B.~Sapp, M.~Bansal, and D.~Anguelov, ``Multipath: {M}ultiple
  probabilistic anchor trajectory hypotheses for behavior prediction,''
  \emph{arXiv preprint arXiv:1910.05449}, 2019.

\bibitem{koskela1996time}
\BIBentryALTinterwordspacing
T.~{Koskela}, M.~{Lehtokangas}, J.~{Saarinen}, and K.~{Kaski}, ``Time series
  prediction with multilayer perception, fir and elman neural networks,'' 1996.
  [Online]. Available: \url{https://academic.microsoft.com/paper/1558675754}
\BIBentrySTDinterwordspacing

\bibitem{colyar2007us}
J.~Colyar and J.~Halkias, ``Us highway 101 dataset,'' \emph{Federal Highway
  Administration (FHWA), Tech. Rep. FHWA-HRT-07-030}, 2007.

\bibitem{huang2018apolloscape}
X.~Huang, X.~Cheng, Q.~Geng, B.~Cao, D.~Zhou, P.~Wang, Y.~Lin, and R.~Yang,
  ``The {A}pollo{S}cape dataset for autonomous driving,'' in \emph{CVPRW},
  2018, pp. 954--960.

\bibitem{chang2019argoverse}
M.-F. Chang, J.~Lambert, P.~Sangkloy, J.~Singh, S.~Bak, A.~Hartnett, D.~Wang,
  P.~Carr, S.~Lucey, D.~Ramanan, \emph{et~al.}, ``Argoverse: 3{D} tracking and
  forecasting with rich maps,'' in \emph{CVPR}, 2019, pp. 8748--8757.

\bibitem{zhu2019starnet}
Y.~Zhu, D.~Qian, D.~Ren, and H.~Xia, ``Star{N}et: {Pedestrian} trajectory
  prediction using deep neural network in star topology,'' \emph{arXiv preprint
  arXiv:1906.01797}, 2019.

\bibitem{lee2017desire}
N.~Lee, W.~Choi, P.~Vernaza, C.~B. Choy, P.~H. Torr, and M.~Chandraker,
  ``Desire: {D}istant future prediction in dynamic scenes with interacting
  agents,'' in \emph{CVPR}, 2017, pp. 336--345.

\bibitem{deo2018would}
N.~Deo, A.~Rangesh, and M.~M. Trivedi, ``How would surround vehicles move? {A}
  unified framework for maneuver classification and motion prediction,''
  \emph{IEEE Transactions on Intelligent Vehicles}, vol.~3, no.~2, pp.
  129--140, 2018.

\bibitem{kuefler2017imitating}
A.~Kuefler, J.~Morton, T.~Wheeler, and M.~Kochenderfer, ``Imitating driver
  behavior with generative adversarial networks,'' in \emph{IV}.\hskip 1em plus
  0.5em minus 0.4em\relax IEEE, 2017, pp. 204--211.

\end{thebibliography}
